\pdfoutput=1
\documentclass{article}
\usepackage{authblk}
\usepackage{bm}
\usepackage{booktabs}
\usepackage{float}
\usepackage[hang,flushmargin]{footmisc}
\usepackage{makecell}
\usepackage{microtype}
\usepackage{multirow}
\usepackage{paralist}
\usepackage{spconf,amsmath,graphicx}
\usepackage{todonotes}
\usepackage{xspace}

\newcommand{\eg}{e.g.,\xspace}

\newcommand{\ie}{i.e.,\xspace}

\makeatletter
\renewcommand\AB@affilsepx{, \protect\Affilfont}
\makeatother

\title{\large \bf Natural TTS Synthesis By Conditioning WaveNet On Mel Spectrogram Predictions}

\author[1]{Jonathan Shen}
\author[1]{Ruoming Pang}
\author[1]{Ron J. Weiss}
\author[1]{Mike Schuster}
\author[1]{Navdeep Jaitly}
\author[2]{Zongheng Yang\thanks{Work done while at Google.}}
\author[1]{Zhifeng Chen}
\author[1]{Yu Zhang}
\author[1]{Yuxuan Wang}
\author[1]{RJ Skerry-Ryan}
\author[1]{Rif A. Saurous}
\author[1]{Yannis Agiomyrgiannakis}
\author[1]{Yonghui Wu}
\affil[1]{Google, Inc.}
\affil[2]{University of California, Berkeley}
\affil[ ]{\texttt{\{jonathanasdf,rpang,yonghui\}@google.com}}
\date{}

\begin{document}
\ninept
\maketitle

\begin{abstract}
This paper describes Tacotron~2,
a neural network architecture for speech synthesis directly
from text. The system is composed of a recurrent sequence-to-sequence feature
prediction network that maps character embeddings to
mel-scale spectrograms, followed by a modified WaveNet model acting as a vocoder
to synthesize time-domain waveforms from those spectrograms. Our model achieves
a mean opinion score (MOS) of $4.53$ comparable to a MOS of $4.58$ for
professionally recorded speech.
To validate our design choices, we present ablation studies of key components of
our system and evaluate the impact of using mel spectrograms as the conditioning
input to WaveNet instead of linguistic, duration, and $F_0$ features. We further
show that using this compact acoustic intermediate representation allows
for a significant reduction in the size of the WaveNet architecture.
\end{abstract}
%

\section{Introduction}
\label{sec:intro}

Generating natural speech from text (text-to-speech synthesis, TTS) remains a
challenging task despite decades of investigation \cite{Taylor:2009:TS:1592988}.
Over time, different techniques have dominated the field.
Concatenative synthesis with unit selection, the process of stitching small
units of pre-recorded waveforms together
\cite{Hunt:1996:USC:1256383.1256532,cstr:unitsel97} was the state-of-the-art for
many years.
Statistical parametric speech synthesis
\cite{Tokuda00speechparameter,Black07statisticalparametric,40837,tokuda2013speech},
which directly generates smooth trajectories of speech features to be
synthesized by a vocoder, followed, solving many of the issues that
concatenative synthesis had with boundary artifacts. However, the audio produced
by these systems often sounds muffled and unnatural compared to human speech.

WaveNet \cite{45774}, a generative model of time domain
waveforms, produces audio quality that begins to rival that of real human
speech and is already used in some complete TTS systems
\cite{DBLP:journals/corr/ArikCCDGKLMRSS17,DBLP:journals/corr/ArikDGMPPRZ17,2017arXiv171007654P}.
The inputs to WaveNet (linguistic features, predicted log fundamental
frequency ($F_0$), and phoneme durations), however, require significant domain
expertise to produce, involving elaborate text-analysis systems as well as a
robust lexicon (pronunciation guide).

Tacotron \cite{46150}, a sequence-to-sequence architecture
\cite{conf/nips/SutskeverVL14} for producing magnitude spectrograms from
a sequence of characters, simplifies the traditional speech synthesis pipeline
by replacing the production of these linguistic and acoustic features with a
single neural network trained from data alone. To vocode the resulting magnitude
spectrograms, Tacotron uses the Griffin-Lim algorithm
\cite{Griffin84signalestimation} for phase estimation, followed by an inverse
short-time Fourier transform. As the authors note, this was simply a placeholder
for future neural vocoder approaches, as Griffin-Lim produces characteristic
artifacts and lower audio quality than approaches like WaveNet.

In this paper, we describe a unified, entirely neural approach to speech
synthesis that combines the best of the previous approaches: a
sequence-to-sequence Tacotron-style model \cite{46150} that generates mel
spectrograms, followed by a modified WaveNet vocoder
\cite{DBLP:journals/corr/ArikDGMPPRZ17,tamamori2017speaker}. Trained directly
on normalized character sequences and corresponding speech waveforms, our
model learns to synthesize natural sounding speech that is difficult to
distinguish from real human speech.

Deep Voice 3 \cite{2017arXiv171007654P} describes a similar approach. However,
unlike our system, its naturalness has not been shown to rival that of
human speech.
Char2Wav \cite{Sotelo2017Char2wavES} describes yet another similar approach to
end-to-end TTS using a neural vocoder. However, they use different intermediate
representations (traditional vocoder features) and their model architecture
differs significantly.


\section{Model Architecture}
\label{sec:arch}

\begin{figure}[t!]
\centering
\includegraphics[width=0.98\columnwidth]{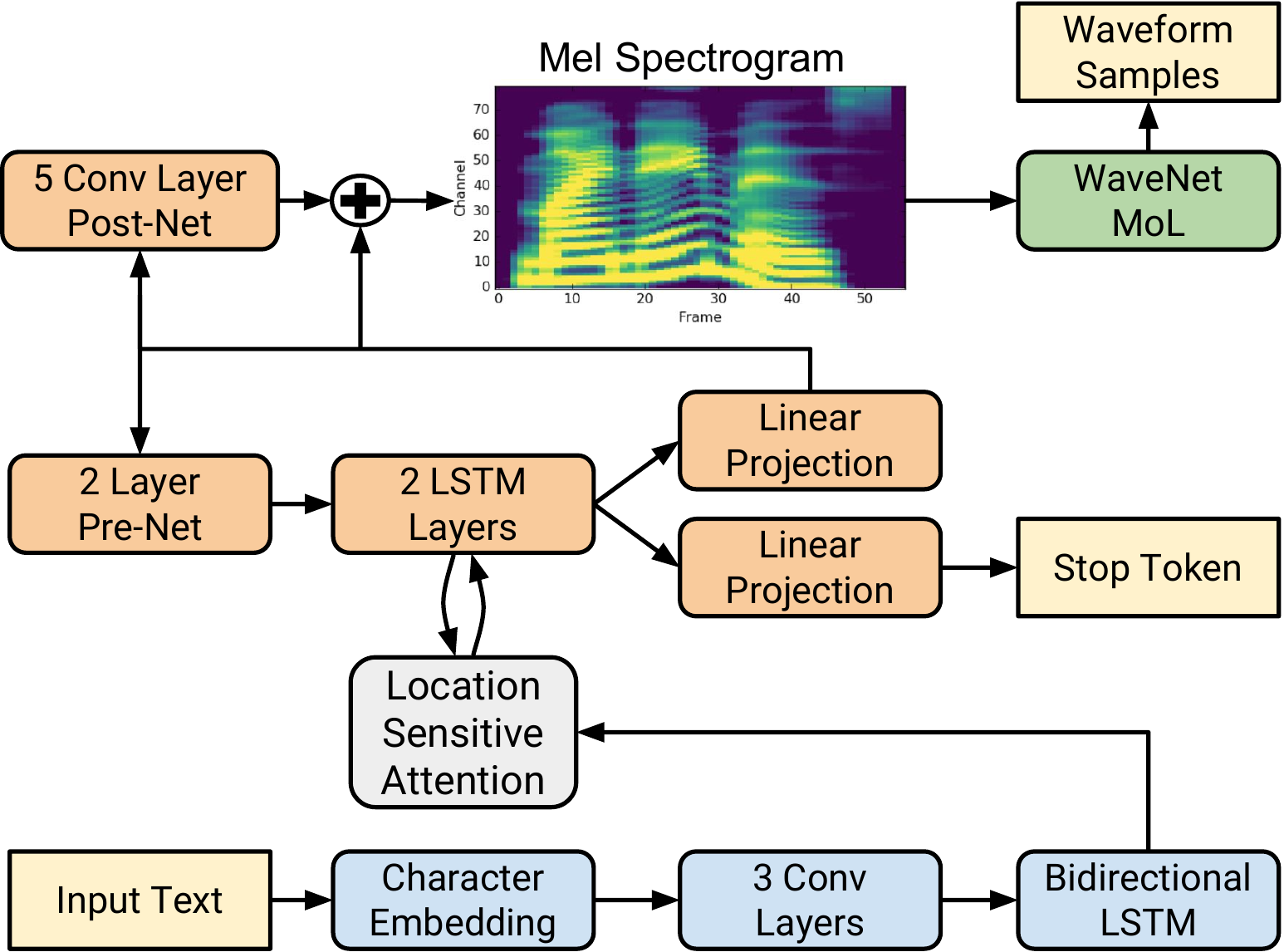}
\caption{Block diagram of the Tacotron 2 system architecture.}
\label{fig:TTSArchitecture}
\end{figure}

Our proposed system consists of two components, shown in Figure~\ref{fig:TTSArchitecture}:
\begin{inparaenum}[(1)]
  \item a recurrent sequence-to-sequence feature prediction network with
    attention which predicts a sequence of mel spectrogram frames from an
    input character sequence, and
  \item a modified version of WaveNet which generates time-domain waveform samples
    conditioned on the predicted mel spectrogram frames.
\end{inparaenum}

\subsection{Intermediate Feature Representation}

In this work we choose a low-level acoustic representation: mel-frequency
spectrograms, to bridge the two components. Using a representation
that is easily computed from time-domain waveforms allows us to train the two
components separately. This representation is also smoother than waveform
samples and is easier to train using a squared error loss because it is
invariant to phase within each frame.

A mel-frequency spectrogram is related to the linear-frequency spectrogram, \ie
the short-time Fourier transform (STFT) magnitude. It is obtained by applying a
nonlinear transform to the frequency axis of the STFT, inspired by measured
responses from the human auditory system, and summarizes the frequency content
with fewer dimensions.
Using such an auditory frequency scale has the effect of emphasizing details in
lower frequencies, which are critical to speech intelligibility, while
de-emphasizing high frequency details, which are dominated by fricatives and
other noise bursts and generally do not need to be modeled with high fidelity.
Because of these properties, features derived from the mel scale have
been used as an underlying representation for speech recognition for
many decades \cite{davis:mel}.

While linear spectrograms discard phase information (and are therefore lossy),
algorithms such as Griffin-Lim \cite{Griffin84signalestimation} are capable of
estimating this discarded information, which enables time-domain conversion
via the inverse short-time Fourier transform. Mel spectrograms discard even more
information, presenting a challenging inverse problem.
However, in comparison to the linguistic and acoustic features used in
WaveNet, the mel spectrogram is a simpler, lower-level acoustic
representation of audio signals. It should therefore be straightforward for a
similar WaveNet model conditioned on mel spectrograms to generate audio,
essentially as a neural vocoder.
%
%
Indeed, we will show that it is possible to generate high quality audio from mel
spectrograms using a modified WaveNet architecture.

\subsection{Spectrogram Prediction Network}
\label{ssec:c2f}


%
As in Tacotron, mel spectrograms are computed
through a short-time Fourier transform (STFT) using a 50~ms frame size, 12.5~ms
frame hop, and a Hann window function. We experimented with a 5~ms frame hop to
match the frequency of the conditioning inputs in the original WaveNet, but
the corresponding increase in temporal resolution resulted in significantly more
pronunciation issues.

%
We transform the STFT magnitude to the mel scale using an 80 channel
mel filterbank spanning 125~Hz to 7.6~kHz, followed by log dynamic
range compression.
Prior to log compression, the filterbank output magnitudes are clipped to a
minimum value of 0.01 in order to limit dynamic range in the logarithmic domain.

The network is composed of an encoder and a decoder with attention.
The encoder converts a character sequence into a hidden
feature representation which the decoder consumes to predict a
spectrogram.
Input characters are represented using a learned 512-dimensional character
embedding, which are passed through
a stack of 3 convolutional layers each containing 512 filters with shape
$5\times1$, \ie where each filter spans 5 characters, followed by batch
normalization \cite{ioffe2015batch} and ReLU activations.
As in Tacotron, these convolutional layers model longer-term
context (\eg $N$-grams) in the input character sequence.
%
The output of the final convolutional layer is passed into a single
bi-directional \cite{Schuster:1997:BRN:2198065.2205129} LSTM
\cite{Hochreiter:1997:LSM:1246443.1246450} layer containing 512 units
(256 in each direction) to generate the encoded features.

The encoder output is consumed by an attention network which
summarizes the full encoded sequence as a fixed-length context vector
for each decoder output step.
We use the location-sensitive attention from
\cite{chorowski2015attention}, which extends the additive attention
mechanism \cite{bahdanau2014neural} to use cumulative attention
weights from previous decoder time steps as an additional feature.
This encourages the model to move forward consistently through the
input, mitigating potential failure modes where some
subsequences are repeated or ignored by the decoder.
%
Attention probabilities are computed after projecting inputs 
and location features to 128-dimensional hidden representations.
Location features are computed using 32 1-D convolution filters of
length 31.

The decoder is an autoregressive recurrent neural network which
predicts a mel spectrogram from the encoded input sequence one
frame at a time.
%
The prediction from the previous time step is first passed through a
small \emph{pre-net} containing 2 fully connected layers of 256 hidden ReLU units.
We found that the pre-net acting as an information bottleneck was essential for
learning attention.
%
The pre-net output and attention context vector are concatenated and
passed through a stack of 2 uni-directional LSTM layers with 1024 units.
The concatenation of the LSTM output and the attention context vector is
projected through a linear transform to predict the target
spectrogram frame.
Finally, the predicted mel spectrogram is passed through a 5-layer convolutional
\emph{post-net} which predicts a residual to add to the prediction to improve the
overall reconstruction.
Each post-net layer is comprised of 512 filters with shape $5\times1$ with
batch normalization, followed by $\tanh$ activations on all but the final layer.

We minimize the summed mean squared error (MSE) from before and after the
post-net to aid convergence.  We also experimented with a log-likelihood loss by
modeling the output distribution with a Mixture Density Network
\cite{Bishop94mixturedensity,Schuster99onsupervised} to avoid assuming
a constant variance over time, but found that these were more difficult to
train and they did not lead to better sounding samples.

In parallel to spectrogram frame prediction, the concatenation of
decoder LSTM output and the attention context
is projected down to a scalar and passed through a sigmoid activation
to predict the probability that the output sequence has completed.
This ``stop token'' prediction is used during inference to allow the model to
dynamically determine when to terminate generation instead of always generating
for a fixed duration.
Specifically, generation completes at the first frame for which this probability
exceeds a threshold of 0.5.

The convolutional layers in the network are regularized using dropout
\cite{srivastava2014dropout} with probability 0.5, and LSTM layers are
regularized using zoneout \cite{krueger2016zoneout} with probability 0.1. In
order to introduce output variation at inference time, dropout with probability
0.5 is applied only to layers in the pre-net of the autoregressive decoder.

In contrast to the original Tacotron, our model uses simpler
building blocks, using vanilla LSTM and convolutional layers in
the encoder and decoder instead of ``CBHG'' stacks and GRU recurrent
layers.
We do not use a ``reduction factor'', \ie each decoder step
corresponds to a single spectrogram frame.

\subsection{WaveNet Vocoder}
\label{ssec:wavenet}

We use a modified version of the WaveNet architecture from \cite{45774} to
invert the mel spectrogram feature representation into time-domain waveform
samples.
As in the original architecture, there are 30 dilated convolution layers,
grouped into 3 dilation cycles, \ie the dilation rate of layer k
($k=0\ldots 29$) is $2^{k\pmod{10}}$.
To work with the 12.5~ms frame hop of the spectrogram frames, only 2 upsampling
layers are used in the conditioning stack instead of 3 layers.

Instead of predicting discretized buckets with a softmax layer,
we follow PixelCNN++ \cite{DBLP:journals/corr/SalimansKCK17} and
Parallel WaveNet \cite{FasterWaveNet} and use a 10-component
mixture of logistic distributions (MoL) to generate 16-bit samples at 24~kHz.
To compute the logistic mixture distribution, the WaveNet stack output is passed
through a ReLU activation followed by a linear projection to predict
parameters (mean, log scale, mixture weight) for each mixture component.
The loss is computed as the negative log-likelihood of the ground truth sample.


\section{Experiments \& Results}
\label{sec:experiments}

\subsection{Training Setup}
Our training process involves first training the feature prediction network
on its own, followed by training a modified WaveNet independently on the outputs
generated by the first network.

To train the feature prediction network, we apply the standard
maximum-likelihood training procedure (feeding in the
correct output instead of the predicted output on the decoder side, also
referred to as {\em teacher-forcing})
with a batch size of 64 on a single GPU. We use the Adam optimizer
\cite{DBLP:journals/corr/KingmaB14} with
$\beta_1=0.9, \beta_2=0.999, \epsilon=10^{-6}$ and a learning rate of
$10^{-3}$ exponentially decaying to $10^{-5}$ starting after 50,000 iterations.
We also apply $L_2$ regularization with weight $10^{-6}$.

We then train our modified WaveNet on the \emph{ground truth-aligned}
predictions of the feature prediction network.
That is, the prediction network is run in teacher-forcing mode,
where each predicted frame is conditioned on the encoded input sequence and the
corresponding previous frame in the ground truth spectrogram. This ensures that
each predicted frame exactly aligns with the target waveform samples.

We train with a batch size of 128 distributed across 32 GPUs
with synchronous updates, using the Adam optimizer with
$\beta_1=0.9, \beta_2=0.999, \epsilon=10^{-8}$ and a fixed learning rate of
$10^{-4}$. It helps quality to average model weights over recent updates.  Therefore
we maintain an exponentially-weighted moving average of the network parameters
over update steps with a decay of 0.9999 -- this version is used for inference
(see also \cite{DBLP:journals/corr/KingmaB14}).
To speed up convergence, we scale the waveform targets by a factor of $127.5$
which brings the initial outputs of the mixture of logistics layer closer to
the eventual distributions.

We train all models on an internal US English dataset\cite{46150}, which
contains 24.6 hours of speech from a single professional female speaker.
All text in our datasets is spelled out. \eg ``16'' is written as ``sixteen'',
\ie our models are all trained on normalized text.

\subsection{Evaluation}

When generating speech in inference mode, the ground truth targets are
not known.  Therefore, the predicted outputs from the previous step
are fed in during decoding, in contrast to the teacher-forcing
configuration used for training.

We randomly selected 100 fixed examples from the test set of our internal
dataset as the evaluation set.
Audio generated on this set are sent to a human rating service similar to
Amazon's Mechanical Turk where each sample is rated by at least 8 raters on a
scale from 1 to 5 with 0.5 point increments, from which a subjective mean opinion score
(MOS) is calculated. Each evaluation is conducted independently from each other,
so the outputs of two different models are not directly compared when
raters assign a score to them.

Note that while instances in the evaluation set never appear in the training
set, there are some recurring patterns and common words between the two sets.
While this could potentially result in an inflated MOS compared to
an evaluation set consisting of sentences generated from random words, using
this set allows us to compare to the ground truth.  Since all the systems
we compare are trained on the same data, relative comparisons are still
meaningful.

Table \ref{tbl:mos_various_systems} shows
a comparison of our method against various prior systems.
In order to better isolate the effect of using mel spectrograms as features,
we compare to a WaveNet conditioned on linguistic features\cite{45774} with similar
modifications to the WaveNet architecture as introduced above. We also compare
to the original
Tacotron that predicts linear spectrograms and uses Griffin-Lim to synthesize
audio, as well as concatenative \cite{gonzalvo2016recent} and parametric
\cite{zen2016fast} baseline systems, both of which have been used in production at Google.
%
We find that the proposed system significantly outpeforms all other TTS systems,
and results in an MOS comparable to that of the ground truth audio.
\footnote[2]{Samples available at https://google.github.io/tacotron/publications/tacotron2.}

\begin{table}[H]
  \centering
  \begin{tabular}{lc}
  \toprule
  System                & MOS \\
  \midrule
  Parametric            & $3.492 \pm 0.096$   \\
  Tacotron (Griffin-Lim)& $4.001 \pm 0.087$   \\
  Concatenative         & $4.166 \pm 0.091$   \\
  WaveNet (Linguistic)  & $4.341 \pm 0.051$   \\
  Ground truth          & $4.582 \pm 0.053$   \\
  \midrule
  Tacotron~2 (this paper)& \bm{$4.526 \pm 0.066$}   \\
  \bottomrule
  \end{tabular}
\caption{Mean Opinion Score (MOS) evaluations with 95\% confidence intervals
computed from the t-distribution for various systems.}
\label{tbl:mos_various_systems}
\end{table}

\begin{figure}[H]
\centering
\includegraphics[scale=0.35]{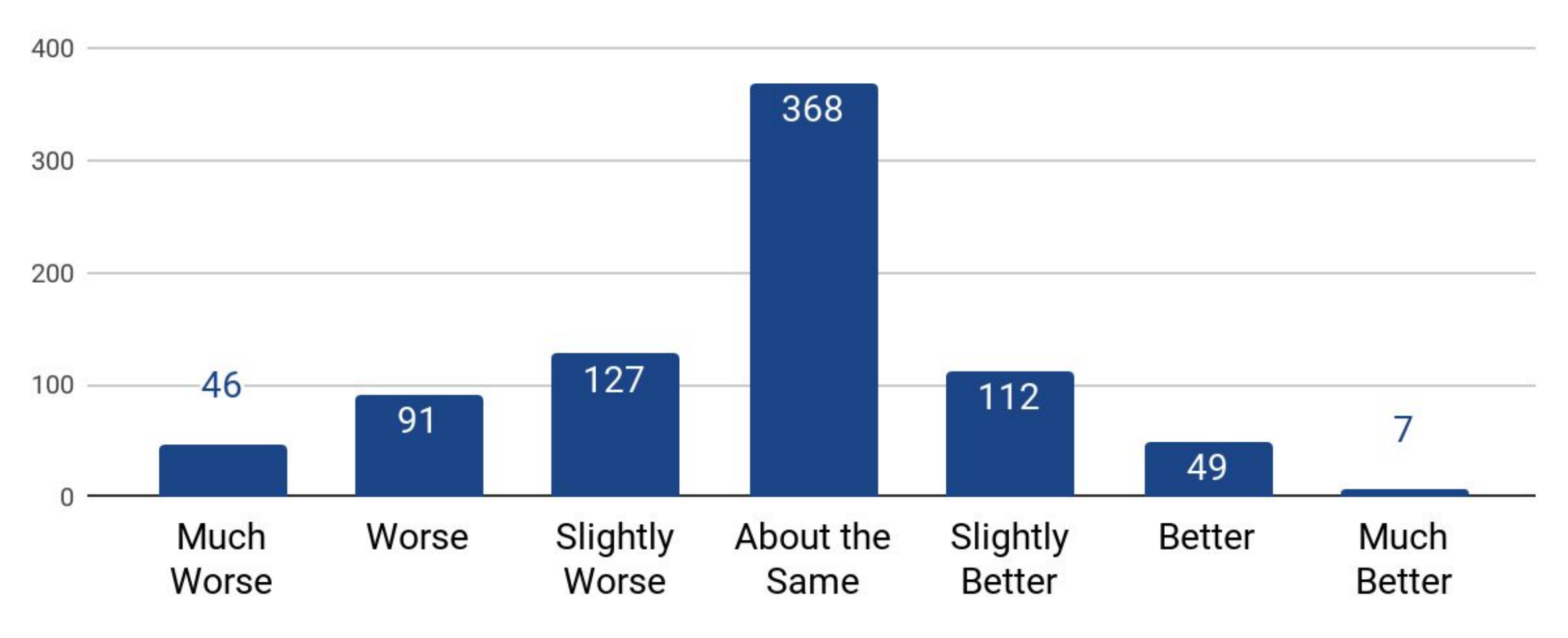}
\caption{Synthesized vs. ground truth: 800 ratings on 100 items.}
\label{fig:sxs_breakdown}
\end{figure}

We also conduct a side-by-side evaluation between audio synthesized by
our system and the ground truth.
For each pair of utterances, raters are asked to give a score ranging from -3
(synthesized much worse than ground truth) to 3 (synthesized much better than
ground truth).
The overall mean score of $-0.270 \pm 0.155$ shows that raters have a small but
statistically significant preference towards ground truth over our results.
See Figure~\ref{fig:sxs_breakdown} for a detailed breakdown.
The comments from raters indicate that occasional mispronunciation by our
system is the primary reason for this preference.

We ran a separate rating experiment on the custom 100-sentence test set from
Appendix E of \cite{2017arXiv171007654P}, obtaining a MOS of 4.354.
In a manual analysis of the error modes of our system, counting errors in
each category independently, 0 sentences contained repeated words,
6 contained mispronunciations, 1 contained skipped words, and 23 were
subjectively decided to contain unnatural prosody, such as emphasis on the
wrong syllables or words, or unnatural pitch. End-point prediction failed in a
single case, on the input sentence containing the most characters.
These results show that while our system is able to reliably attend to the
entire input, there is still room for improvement in prosody modeling.

Finally, we evaluate samples generated from 37 news headlines to test the
generalization ability of our system to out-of-domain text. On this task, our
model receives a MOS of $4.148 \pm 0.124$ while WaveNet conditioned on
linguistic features receives a MOS of $4.137 \pm 0.128$.
A side-by-side evaluation comparing the output of these systems also
shows a virtual tie -- a statistically insignificant preference towards our
results by $0.142 \pm 0.338$. Examination of rater comments shows that our
neural system tends to generate speech that feels more natural and human-like,
but it sometimes runs into pronunciation difficulties, \eg when
handling names. This result points to a challenge for end-to-end
approaches -- they require training on data that cover intended usage.

\subsection{Ablation Studies}
\label{sec:ablation}

\subsubsection{Predicted Features versus Ground Truth}
\label{ssec:gtfeats}

While the two components of our model were trained separately, the WaveNet component
depends on the predicted features for training. An alternative
is to train WaveNet independently on mel spectrograms extracted from ground
truth audio. We explore this in Table~\ref{tbl:pred_gt}.

\begin{table}[H]
  \centering
  \begin{tabular}{lcc}
    \toprule
                 & \multicolumn{2}{c}{Synthesis}\\
    Training     & Predicted         & Ground truth \\
    \midrule
    Predicted    & $4.526 \pm 0.066$ & $4.449 \pm 0.060$ \\
    Ground truth & $4.362 \pm 0.066$ & $4.522 \pm 0.055$ \\
    \bottomrule
  \end{tabular}
\caption{Comparison of evaluated MOS for our system when WaveNet trained on
predicted/ground truth mel spectrograms are made to synthesize from
predicted/ground truth mel spectrograms.}
\label{tbl:pred_gt}
\end{table}

As expected, the best performance is obtained when the features used for
training match those used for inference. However, when trained on
ground truth features and made to synthesize from
predicted features, the result is worse than the opposite.
This is due to the tendency of the predicted spectrograms to be oversmoothed and less
detailed than the ground truth -- a consequence of the squared error loss
optimized by the feature prediction network. When trained on ground truth
spectrograms, the network does not learn to generate high quality speech
waveforms from oversmoothed features.

\subsubsection{Linear Spectrograms}
\label{ssec:linear}

Instead of predicting mel spectrograms, we experiment with training
to predict linear-frequency spectrograms instead, making it
possible to invert the spectrogram using Griffin-Lim.

\begin{table}[H]
  \centering
  \begin{tabular}{lc}
  \toprule
  System                        & MOS \\
  \midrule
  Tacotron~2 (Linear + G-L)     & $3.944 \pm 0.091$   \\
  Tacotron~2 (Linear + WaveNet) & $4.510 \pm 0.054$   \\
  Tacotron~2 (Mel + WaveNet)    & $\bm{4.526 \pm 0.066}$ \\
  \bottomrule
  \end{tabular}
\caption{Comparison of evaluated MOS for Griffin-Lim vs. WaveNet as a vocoder,
and using 1,025-dimensional linear spectrograms vs. 80-dimensional
mel spectrograms as conditioning inputs to WaveNet.}
\end{table}

As noted in \cite{DBLP:journals/corr/ArikDGMPPRZ17}, WaveNet produces much
higher quality audio compared to Griffin-Lim. However, there is not much
difference between the use of linear-scale or mel-scale spectrograms. As such,
the use of mel spectrograms seems to be a strictly better choice since it is a
more compact representation. It would be interesting to explore the
trade-off between the number of mel frequency bins versus audio quality
in future work.

\subsubsection{Post-Processing Network}
\label{ssec:postedit}

Since it is not possible to use the information of predicted future frames
before they have been decoded, we use a convolutional post-processing network
to incorporate past and future frames after decoding to improve the feature
predictions. However, because WaveNet already contains convolutional layers,
one may wonder if the post-net is still necessary when WaveNet is used as the
vocoder. To answer this
question, we compared our model with and without the post-net, and found that
without it, our model only obtains a MOS score of $4.429 \pm 0.071$, compared to
$4.526 \pm 0.066$ with it, meaning that empirically the post-net is still an
important part of the network design.

\subsubsection{Simplifying WaveNet}
\label{ssec:simplifywavenet}

A defining feature of WaveNet is its use of dilated convolution to increase
the receptive field exponentially with the number of layers.
We evaluate models with varying receptive field sizes and number of
layers to test our hypothesis that a shallow network with a small receptive
field may solve the problem satisfactorily since mel spectrograms are a much
closer representation of the waveform than linguistic features and already
capture long-term dependencies across frames.

As shown in Table \ref{tbl:wavenets}, we find that our model can generate
high-quality audio using as few as 12 layers with a receptive field of
10.5~ms, compared to 30 layers and 256~ms in the baseline model. These
results confirm the observations in \cite{DBLP:journals/corr/ArikCCDGKLMRSS17}
that a large receptive field size is not an essential factor for audio
quality. However, we hypothesize that it is the choice to condition on mel
spectrograms that allows this reduction in complexity.

On the other hand, if we eliminate dilated convolutions altogether, the
receptive field becomes two orders of magnitude smaller than the baseline and
the quality degrades significantly even though the stack is as deep as the
baseline model.
This indicates that the model requires sufficient context at the time scale of
waveform samples in order to generate high quality sound.

\begin{table}[H]
  \centering
  \begin{tabular}{ccccc}
  \toprule
  \makecell{Total\\layers} & \makecell{Num\\cycles} &
  \makecell{Dilation\\cycle size} & \makecell{Receptive field\\(samples / ms)} &
  MOS \\
  \midrule
  30 & 3  & 10 & 6,139 / 255.8 & $4.526 \pm 0.066$   \\
  24 & 4 &  6 & 505 / 21.0 & $4.547 \pm 0.056$   \\
  12 & 2 &  6 & 253 / 10.5 & $4.481 \pm 0.059$   \\
  30 & 30 &  1 & 61 / 2.5 & $3.930 \pm 0.076$ \\
  \bottomrule
  \end{tabular}
\caption{WaveNet with various layer and receptive field sizes.}
\label{tbl:wavenets}
\end{table}


\section{Conclusion}
\label{sec:conclusion}

This paper describes Tacotron~2, a fully neural
TTS system that combines a sequence-to-sequence recurrent network with attention
to predicts mel spectrograms with a modified WaveNet vocoder.  The resulting system synthesizes
speech with Tacotron-level prosody and WaveNet-level audio quality. This system
can be trained directly from data without relying on complex feature
engineering, and achieves state-of-the-art sound quality close to that of
natural human speech.


\section{Acknowledgments}
The authors thank Jan Chorowski, Samy Bengio, A{\"a}ron van den Oord, and the
  WaveNet and Machine Hearing teams for their helpful discussions and advice, as
  well as Heiga Zen and the Google TTS team for their feedback and assistance
  with running evaluations.
  The authors are also grateful to the very thorough reviewers.
\bibliographystyle{IEEEbib}
\bibliography{ms}

\end{document}